\documentclass[12pt,a4paper]{article}
\usepackage{authblk}
\usepackage{graphicx}
\usepackage[colorlinks=true, allcolors=blue]{hyperref}
\usepackage[utf8]{inputenc}
\usepackage[a4paper,top=2cm,bottom=2cm,left=2cm,right=2cm,marginparwidth=1.75cm]{geometry}
\usepackage{amsmath}
\usepackage{amssymb}
\usepackage{wrapfig}
\usepackage{algorithm}
\usepackage{algpseudocode}
\usepackage{graphicx}
\usepackage{multirow}
\usepackage{caption}
\usepackage{booktabs}

\newcommand{\ourapp}{Compositional Neural Operators}
\newcommand{\abrev}{CompNO}

\title{Compositional Neural Operators for Multi-Dimensional Fluid Dynamics}
\author[a]{Hamda HMIDA}

\author[b]{Hsiu-Wen CHANG JOLY}
\author[a]{Youssef MESRI \thanks{Corresponding Author: youssef.mesri@minesparis.psl.eu}}

\affil[a]{Mines Paris - PSL University,
Centre for Material Forming (CEMEF)}
\affil[b]{Mines Paris - PSL University,
Centre for Robotics (CAOR)}
 
\date{March 2026}

\begin{document}

\maketitle

\begin{abstract}
\noindent Partial differential equations (PDEs) govern diverse physical phenomena, yet high-fidelity numerical solutions are computationally expensive and Machine Learning approaches lack generalization. While Scientific Foundation Models (SFMs) aim to provide universal surrogates, typical encoding-decoding approaches suffer from high pretraining costs and limited interpretability. In this paper, we propose \ourapp~(\abrev)\footnote{Code and Data  will be made available on request.} for 2D systems, a framework that decomposes complex PDEs into a library of Foundation Blocks. Each block is a specialized Neural Operator pretrained on elementary physics. This modular library contains convection, diffusion, and nonlinear convection blocks as well as a Poisson Solver, enabling the framework to address the pressure-velocity coupling. These experts are assembled via an Adaptation Block featuring an Aggregator. This aggregator learns nonlinear interactions by minimizing data loss and physics-based residuals driven from governing equations. The proposed approach has been evaluated on the Convection-Diffusion equation, the Burgers’ equation, and the Incompressible Navier-Stokes equation. Our results demonstrate that learning from elementary operators significantly improves adaptability, enhances model interpretability and facilitates the reuse of pretrained blocks when adapting to new physical systems.\\
    \textbf{Key words.} Scientific foundation models; Neural operators; PDE simulation; Computational fluid dynamics; Fourier neural operator, Physics-Informed Neural Networks.
\end{abstract}

\section{Introduction}
Solving partial differential equations (PDEs) is a cornerstone of modeling and simulating complex phenomena in science and engineering. In fields such as computational fluid dynamics (CFD), PDEs serve as the primary framework for describing fluid behavior, relating variables like velocity and pressure to the system's evolution over time. Traditional numerical methods, including finite difference, finite volume, and finite element methods, provide high-fidelity results but often require significant computational resources. This cost becomes particularly prohibitive when simulations must be repeated across wide parameter variations or for real-time applications. Recently, machine/deep learning approaches have emerged as promising alternatives. Some discrete data-driven methods represent PDE solutions directly with neural networks tailored to specific equations and discretization. Early work leveraged Convolutions \cite{Zhu_2018, Bhatnagar_2019}, Residual \cite{TAYLOR2023115850}, and Graph \cite{pelissier2024graph} network architectures to encode spatial information and evolve solutions over time, while physics-informed frameworks (PINNs \cite{RAISSI2019686, yang2022learning}, hPINN \cite{doi:10.1142/S0129183123500821}, PIKANs \cite{Wang_2025}) enabled unsupervised and semi-supervised recovery of single-instance solutions. The problem with these neural solvers is their limited ability to handle discretization effectively as well as their dependency on system configuration. Other continuous frameworks like neural operators \cite{kovachki2023neural} and DeepONet \cite{lu_deeponet_2021} can learn mappings between function spaces by approximating operators instead of data distribution. DeepONet \cite{lu_deeponet_2021} employs a dual network design to map infinite-dimensional functions, while the Fourier Neural Operator (FNO) \cite{li2020fourier} leverages the fast Fourier transform to achieve discretization invariance. Some extensions, like GINO \cite{li2023geometry}, were proposed to adapt it to complex geometries, while other work, such as PI-DeepONet \cite{wang2021learning} and PFNO \cite{yu2024parametric}, were trying to enhance the performance of neural operators, but they still face challenges related to limited generalization abilities.\\

\noindent Recent advancements have shifted focus toward the development of Scientific Foundation Models (SFMs). These are data-driven models trained on a broad distribution of physical systems, designed to exhibit wide generalization capabilities across different scientific problems and conditions without requiring retraining from scratch \cite{choi2025definingfoundationmodelscomputational}. SFMs \cite{totounferoush2025pavingwayscientificfoundation, MENON2026108567} serve as reusable bases that reduce the need for specific solvers for every unique PDE type. Researchers have explored several avenues for these models, including using language models to process symbolic PDE forms and utilizing simulation data to predict future states.\\

\noindent Despite their promise, current SFMs face notable challenges, such as the high computational expense of pretraining on massive, heterogeneous physics datasets and a lack of physical interpretability in their architectures. In this work, we present \ourapp~ (\abrev), a foundation model approach that aims to address these limitations through the principle of compositionality. Building upon the modular framework introduced by \cite{app16020972}, which demonstrated efficacy in one-dimensional settings, our work significantly extends the scope and utility of compositional neural operators in three key ways. First, we transition from one-dimensional setups to complex two-dimensional systems, specifically targeting the 2D Incompressible Navier-Stokes (INS) equations. Second, we introduce Physics-Informed Aggregation within the Adaptation Blocks to ensure that the assembly of Foundation Blocks remains consistent with underlying physical constraints. Finally, by constructing a library of specialized blocks pretrained on fundamental differential operators, \abrev~ demonstrates superior performance over well-established methods. This modular strategy improves data efficiency, enhances physical interpretability, and enables robust generalization across diverse, high-dimensional physical regimes.
\section{Related Work}
To develop foundation models capable of solving diverse PDE systems, recent work has explored two primary paradigms. The first paradigm leverages language models that condition solution prediction on the symbolic form of the governing PDE, treating mathematical expressions as structured sequences analogous to text. Representative examples include PROSE \cite{liu2024prose} and PROSE-FD \cite{liu2024prose_fd}, which employ multimodal transformer architectures to jointly encode symbolic PDE descriptions and numerical solution data, enabling zero-shot or few-shot generalization across equation families. A related approach \cite{yang2025fine} fine-tunes large pretrained language models to incorporate textual or symbolic representations of PDEs as conditioning context for operator learning. Similarly, Unified Pretrained Solvers (UPS) \cite{shen2024ups} integrate symbolic and numerical modalities through large-scale pretraining, yielding a single model that transfers effectively across multiple PDE classes with limited task-specific data.\\

\noindent The second paradigm dispenses with explicit symbolic supervision and relies exclusively on simulation data, learning to predict future states from previous observations. Within this setting, VICON \cite{cao2024vicon} exploits in-context learning to infer solution operators directly from example trajectories, enabling rapid adaptation to new PDE instances. BCAT \cite{liu2025bcatblockcausaltransformer} introduces block-causal transformers for autoregressive prediction of high-dimensional fluid states, while Poseidon \cite{herde2024poseidon} employs multiscale operator transformers pretrained on fluid dynamics datasets to achieve strong cross-equation generalization. Multiple Physics Pretraining (MPP) \cite{mccabe2024multiple} further extends this approach by training a single model across datasets drawn from multiple PDE systems, encouraging the emergence of shared physical representations. Furthermore, PhysiX \cite{nguyen2025physixfoundationmodelphysics} adopts large-scale autoregressive modeling inspired by video transformers, demonstrating that foundation models trained purely on spatiotemporal simulation data can generalize across diverse physical systems without access to explicit governing equations.\\

\noindent Despite these advances, existing foundation models for PDEs exhibit several limitations. Many approaches require large volumes of expensive simulation data and substantial computational resources for pretraining, and often rely on opaque latent representations that limit interpretability. In contrast, \abrev~ \cite{app16020972} overcomes these challenges by eschewing a monolithic latent space for heterogeneous physical systems. Instead, \abrev~ is first pretrained to learn dynamic representations of common differential operators. It is then fine-tuned to solve specific PDE problems by routing and composing the operators that govern the target system.

\section{Methodology}
\subsection{Problem Setting}
We consider parametric partial differential equations defined in a domain $D \subset {\rm I\!R}^n$ and $P \subset {\rm I\!R}^p$ is the parameter space where $p$ is the number of parameters of this PDE. We define a parametric operator $H$ as a time-stepping solver. Given the state of the system $u_t \in D$ at time $t$ and a set of physical parameters $\gamma \in P$, the model is designed to predict the incremental change $\delta u_t$ over a discrete time step $\Delta t$:
\begin{equation}
    \delta u_t = H(u_t, \gamma )
\end{equation}
then the state at the subsequent time step is determined by the update rule:
\begin{equation}
    u_{t+1} = u_t + \delta u_t
\end{equation}
This formulation allows the model to focus on learning the underlying dynamics rather than the absolute magnitude of the field. 

\noindent We approximate the operator $H$ using a neural operator  $H_{\theta}$, where $\theta \in \Theta$ is the space of all trainable parameters in the neural network. Formally, the neural operator represents a mapping between function spaces:
\begin{equation*}
    H_\theta : D \times P \rightarrow D 
\end{equation*}
The theoretical objective is to identify the optimal parameter set $\theta^*$ in the parameter space $\Theta$ that minimizes the risk across the entire distribution of initial conditions and physical parameters:
\begin{equation} 
    \theta^* = \arg \min_{\theta \in \Theta} \mathbb{E} \left[ \mathcal{L}(\mathcal{H}_\theta(u_t, \gamma), \delta u_t^{Truth}) \right] 
\end{equation}
However, in practice, we have access to a finite dataset of $N$ observations. Therefore, the training process yields an estimated parameter set $\hat{\theta}$, which is the minima of the empirical loss function:
\begin{equation}
    \hat{\theta} = \arg \min_{\theta \in \Theta} \frac{1}{N} \sum_{i=1}^{N} \mathcal{L}(\mathcal{H}_\theta(u_t^{(i)}, \gamma^{(i)}), \delta u_t^{(i), Truth})
\end{equation}
While $\hat{\theta}$ is the optimal solution for the provided observations, it remains an approximation of the true optimal $\theta^*$. The convergence of $\hat{\theta}$ to $\theta^*$ is governed by the diversity of the training data.

\noindent Then, the ultimate goal of $H_{\hat{\theta}}$ is to generalize from a known sequence of time steps to predict the long-term evolution of the system. Given initial observation up to time $T_0$:
\begin{equation*}
    \{ u_{i} \text{  } | \text{  } 0\leq i \leq T_0 \}
\end{equation*}
the model performs an autoregressive rollout to predict the next $T$ time steps.
\begin{equation} 
    \hat{\mathbf{U}} = \{ \hat{u}_{i+1} \mid \hat{u}_{i+1} = \hat{u}_{i} + \mathcal{H}_{\hat{\theta}}(\hat{u}_{i}, \gamma) \text{ for } T_0 < i < T_0 + T \} 
\end{equation}

\noindent In this iterative process, each prediction $\hat{u}_{i+1}$ is fed back into the model as the input for the following inference cycle. This approach tests the model's ability to minimize error accumulation and maintain physical consistency over extended temporal horizons.
\subsection{General Formulation of \abrev}
\begin{figure}[h]
    \centering
    \includegraphics[width=\linewidth]{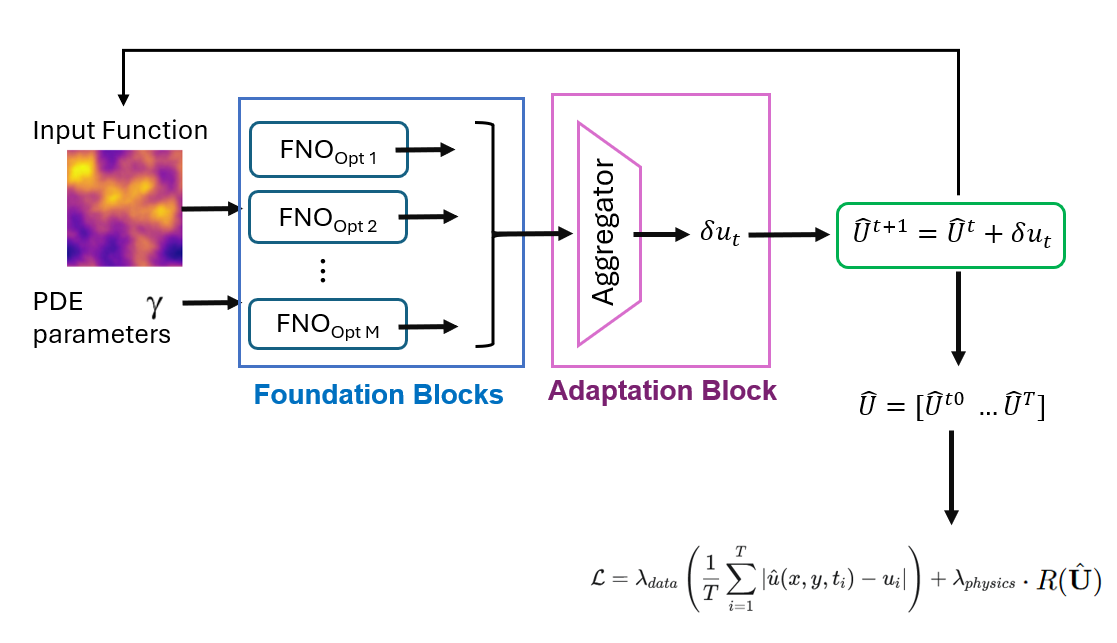}
    \caption{\abrev~ model overview: The input consists of the initial function state $\hat U^t$ with the physical parameter vector $\gamma$. The Foundation Blocks are pretrained Neural Operators that independently predict the time-evolution corresponding to specific elementary operators. The Aggregator is a neural network that combines their embeddings. The prediction $\hat U^{t+1}$ will be used as input for the next time step prediction. The training process incorporates data-driven loss as well as physics loss.}
    \label{fig:arch}
\end{figure}

\ourapp~is built on the principle of compositionality, which states that the behavior of complex physical systems emerges from the interactions of their individual components. This mathematical perspective allows us to translate physical modularity directly into our neural framework through a modular design consisting of Foundation Blocks and an Adaptation Block, as illustrated in Figure \ref{fig:arch}.\\

\noindent The translation of compositionality into our architecture follows a "pre-training assemble fine-tune" paradigm, where Foundation Blocks are pretrained to approximate elementary operators identified in the governing equations. Then, during fine-tuning, the Adaptation Block provides the physical context for the solutions generated by the Foundation Blocks.

\subsection{Foundation Blocks}
In our architecture illustrated in Figure \ref{fig:arch}, the Foundation Blocks are a library consisting of expert modules pretrained to capture the dynamics of elementary physical operators in a 2D domain. Each block is a Parametric Fourier Neural Operator (PFNO), which enables discretization invariance and parametric dependence, or the original Fourier Neural Operator (FNO) if the governing equation lacks external parameters (architecture details in Appendix \ref{fno}).\\

\noindent The blocks are subjected to an autoregressive sequence-to-sequence training to learn the next time step prediction for a fundamental operator $O_i$. We define the pretraining task by the canonical form: 
\begin{equation}
    \frac{\partial \mathbf{u}}{\partial t} + \gamma \cdot O_i( \mathbf{u}) = 0,
\end{equation}
This ensures that each foundation block $\hat{O}_i$ develops an exclusive representation of a unique elementary operator $O_i$. \\

\noindent Once pretraining is complete, the weights of these blocks are frozen, and the final projection layer of each i-th foundation block ($\hat{O}_i$) is removed to supply a high-dimensional embedded representation of the solution $\epsilon_i$, defined as:\\
\begin{equation} 
    \epsilon_i = \hat{O}_i^{emb}(\mathbf{u}_t, \gamma) \in \mathbb{R}^{d_{emb}}
\end{equation}
where $d_{emb}$ is the embedding dimension. This representation $\epsilon_i $ acts as a physical dictionary that captures structural patterns, spatial gradients, and parameter dependencies. By supplying these enriched embeddings rather than raw values, the foundation blocks provide the aggregator with a more informative basis for learning complex multi-physics interactions.

\subsection{Adaptation Blocks}
In our architecture illustrated in Figure \ref{fig:arch}, the Adaptation Block is the task-specific component of the framework designed to assemble the foundation experts into a unified solver for the target PDE. While the Foundation Blocks capture the physical behaviors of the operators separately, the Aggregator $\mathcal{A}$ learns the interactions, couplings, and nonlinearities unique to the system being solved. \\

\noindent The Aggregator $\mathcal{A}$ is a learnable multilayer perceptron (MLP) that processes the concatenated high-dimensional latent embeddings $\epsilon_i $ from the selected Foundation Blocks of each rollout step. We model the one-step incremental prediction as:
\begin{equation} 
    \delta \hat{\mathbf{u}}_t = \mathcal{A}(\epsilon_1 \oplus  \dots \oplus \epsilon_{opt})
\end{equation}
where $\oplus$ denotes the concatenation operator across the feature dimension. By operating in this enriched latent space rather than on raw physical values, the aggregator can exploit the pre-learned structural and gradient information encoded within each $\epsilon_i $. The selection of specific foundation operators is guided by the mathematical structure of the target PDE. In practice, we manually assemble these blocks based on the structure of the target equations.\\

\noindent Mathematically, we view the target PDE as a composite operator $O_{target}$. The Adaptation Block effectively learns a mapping that approximates this composition:
\begin{equation}
    O_{target} = \sum_{i=1}^{opt} \alpha_i  \hat{O}_i^{emb}
\end{equation}
where $\hat{O}_i^{emb}$ are foundation operators involved in the target physics and $\alpha_i$ are the sets of parameters of the aggregator neural network.

\noindent The training of the Aggregator $A$ is driven by an optimization criterion designed to bridge the gap between individual operator dynamics and the full coupled system. The objective function of \abrev~ is a dual-term criterion:
\begin{equation} 
    \mathcal{J}(\alpha) = \mathcal{C}_{FB} + \mathcal{C}_{cross} 
\end{equation}
where:
\begin{itemize}
    \item $\mathcal{C}_{FB}$ represents the direct approximation of the elementary operators within target PDE learned by the Foundation Blocks. In the \abrev~ framework, this term is assumed to be near-zero, as the core physical knowledge of the constituent operators is  accurately captured and transferred by the frozen Foundation Blocks.
    \item $\mathcal{C}_{cross}$ is the interaction term, representing the inner coupling and nonlinear interactions between the operators. Mathematically, this can be viewed as the closure law of the system. While the Foundation Blocks provide the latent $\epsilon_i$, the Aggregator must learn how these fields interfere, compete, or amplify each other. ( More details in Appendix \ref{closure})
\end{itemize}

\noindent To ensure the modeled interaction $\mathcal{C}_{cross}$ remains physically grounded, the aggregation is guided by a Physics-Informed loss function; The aggregator is trained to minimize the residual R derived from the target PDE as well as the data loss, balanced by coefficients $\lambda_{physics}$ and $\lambda_{data}$. The loss function can be defined as shown in the Figure \ref{fig:arch}:
\begin{equation}
    \mathcal{L} = \lambda_{data} \cdot \mathcal{L}_{data} + \lambda_{physics} \cdot R(\hat{\mathbf{U}})
\end{equation}
with $\mathcal{L}_{data}$ is the Mean Absolute Error (MAE) loss function, defined as:\\
\begin{equation}
    MAE = \frac{1}{T} \sum_{t \le T}|u_t-\hat{u}_t|.
\end{equation}
where T is the final time, $t$ represents the time step $\in [0, ..., T]$, $u_t$ is the ground truth target, and $\hat{u}_t$ represents the model's prediction.\\ 

\noindent To solve the case of INS, the Aggregator $\mathcal{A}$ handles the coupling required to satisfy the incompressibility equation inspired by the classical Projection Method \cite{GUERMOND20066011}, which decomposes the velocity update into a convection-diffusion step followed by a pressure-gradient correction needed to satisfy the divergence-free constraint:
\begin{equation}
    \nabla \cdot ( \hat{\mathbf{u}}_t + \delta \hat{\mathbf{u}}_t ) = 0
\end{equation}

\noindent We provide the Aggregator with the latent embeddings from the relevant experts: the nonlinear convection embedding $\epsilon_{conv}$, the diffusion embedding $\epsilon_{diff}$, and the gradient of the pressure embedding $\nabla \epsilon_{p}$ derived from the Poisson Foundation Block. To maintain numerical consistency with the underlying physics, we apply the gradient operator $\nabla$ on the embedded features $\epsilon_p$ using central differences. The Aggregator then approximates the following multi-stage physical process:
\begin{itemize}
    \item Intermediate State Construction: The foundation blocks $\hat{\mathcal{O}}_{conv}$ and $\hat{\mathcal{O}}_{diff}$ provide the latent representation of the intermediate velocity $\mathbf{u}^*$, which accounts for advection and viscosity but is not necessarily divergence-free:
    \begin{equation}
    \mathbf{u}^* \approx \mathbf{u}_t + \mathcal{A}_{sub}(\epsilon_{conv} \oplus \epsilon_{diff})
    \end{equation}
    
    \item Pressure-Gradient Correction: To satisfy the incompressibility constraint, the Aggregator utilizes the pressure information $\epsilon_{p}$ from the Poisson block. In classical CFD, the update follows $\mathbf{u}_{t+1} = \mathbf{u}^* -  \frac{\Delta t}{\rho} \nabla p$. In our framework, the Aggregator learns to perform this projection in the latent space:
    \begin{equation}
    \delta \hat{\mathbf{u}}_t = \mathcal{A} \left( (\epsilon_{conv} \oplus \epsilon_{diff}) \oplus \nabla \epsilon_{p} \right)
    \end{equation}
\end{itemize}

\noindent By providing $\nabla \epsilon_{p}$ as a direct input, the Aggregator is equipped with the necessary geometric information to correct the velocity field. Instead of forcing a single network to learn the entire projection, \abrev~ routes the required physical components through the Aggregator, which then learns the closure law ($\mathcal{C}_{cross}$) that balances momentum and mass conservation.

\section{Results}\label{results}
In this section, we evaluate the performance of \abrev~ in two-dimensional domains, assessing its ability to reconstruct complex fluid dynamics from the knowledge transferred from elementary operators. We specifically investigate the framework's capability to maintain physical consistency quantified via the physics residual loss compared to the monolithic PFNO baseline. All datasets utilized for training and evaluation were generated using high-fidelity numerical methods to serve as the ground-truth physics; for a comprehensive description of data generation methods, the reader is referred to Appendix \ref{data_gen}.
\subsection{Pretraining Performance}
The first stage of the proposed framework involves the independent pretraining of the foundation library. To ensure the robustness of the experts, the Foundation Blocks were trained on a diverse set of parameters and initial conditions, including the Taylor–Green Vortex, Shear Layer, and Isotropic Turbulence. The Foundation Blocks were constructed using four Fourier integral operator layers with the GELU activation function.  The embedding dimension used for these experiments is $d_{emb} = 128$. These blocks are trained using only the MAE loss function.\\

\noindent For the current experiments, we constructed a library of four Foundation Blocks:
\begin{itemize}
    \item Convection Block: Trained on the linear convection equation  with the parameter $\beta$ taking values from the set $\{0.5, 1.0, 1.5, 2.0, 2.5\}$.
    \item Diffusion Block: Trained on the diffusion equation   where the parameter $\nu$ is sampled from the set $\{1, 0.1, 0.01, 0.001\}$.
    \item Nonlinear Convection Block: Trained on the inviscid Burgers' equation  to capture nonlinear convection dynamics.
    \item Poisson Block: Trained on the pressure-Poisson equation to provide the pressure field required for incompressible flows.
\end{itemize}

\noindent Table \ref{tab:pre} reports the MAE values obtained during the pretraining phase. The dataset is split into 80\% of $(128 \times 128)$ grid and 20\% of finer resolution with $(256 \times 256)$ grid. \\
\begin{table}[h]
\centering
\caption{MAE losses for different pretrained models.}
\begin{tabular}{lcc}
\toprule
 Model  & Scalar Training Loss & Vectorial Training Loss  \\
\midrule
 PFNO Convection  & 1 $\times$ $10^{-5}$ & - \\

 PFNO Diffusion  & 2 $\times$ $10^{-5}$ & 8 $\times$  $10^{-5}$ \\

 FNO NL Convection  & 5 $\times$ $10^{-5}$ & 2 $\times$ $10^{-4}$ \\

 FNO Poisson  & - & 2 $\times$ $10^{-5}$ \\
 \bottomrule
\end{tabular}
\label{tab:pre}
\end{table}\\
As shown in Table \ref{tab:pre}, our pretrained blocks achieve a great performance minimizing the MAE in a good way. The low error rates across both scalar and vectorial operators confirm that our Foundation Blocks successfully learn the underlying operators in isolation, providing a rich latent basis for subsequent aggregation.\\

\subsection{Scalar Aggregation}
The scalar track evaluates the model on coupled 2D Convection-Diffusion and Viscous Burgers'. For these experiments, the models are provided with an initial state $T_0=1$ and tasked with an autoregressive rollout of $T=30$ time steps.

\begin{table}[h]
\centering
\caption{Temporal evolution training MAE of loss for different metrics.}
\begin{tabular}{lccccc}
\toprule
\multirow{2}{*}{PDE} & \multirow{2}{*}{Pe / Re} & \multirow{2}{*}{Model} & Trainable & \multicolumn{1}{c}{MAE} & \multirow{2}{*}{Physics Loss} \\
 & &  & Parameters & Velocity Loss & \\
\midrule
\multirow{2}{*}{Convection-Diffusion} & \multirow{2}{*}{$Pe \in [1, 250] $} & \abrev  & 41 K  & 3 $\times$  $10^{-4}$  & 0.05 \\
 && PFNO & 465 K & 9 $\times$  $10^{-3}$  & 24 \\
\midrule
\multirow{2}{*}{Burgers'} & \multirow{2}{*}{$Re \in [1, 100] $} & \abrev & 75 K & 2 $\times$  $10^{-3}$  & 0.6 \\
 && PFNO &  465 K & 0.019 &  19.94 \\
\bottomrule
\end{tabular}\\
\label{tab:model_u}
\end{table}
\noindent As summarized in Table \ref{tab:model_u}, \abrev~ consistently outperforms the monolithic PFNO when trained on the exact same amount of data. A critical observation is the Physics Loss discrepancy; while PFNO achieves reasonable velocity accuracy, its inability to strictly respect the PDE constraints leads to a significantly higher residual error. In contrast, our modular approach preserves the learned dynamics of the frozen experts through Physics-Informed Aggregation, ensuring a physically consistent temporal evolution that adheres more closely to the governing equations.

\subsection{Vectorial Aggregation}
The vectorial track addresses the bi-variable version of the Viscous Burgers' equation and the Incompressible Navier-Stokes equations, where the model must handle coupled velocity components (u,v) with the pressure for the INS case. The models are provided with a sequence $T_0=10$ and perform a rollout of $T=50$ time steps.
\begin{table}[h]
\centering
\caption{Temporal evolution training loss for different metrics.}
\begin{tabular}{lccccc}
\toprule
\multirow{2}{*}{PDE}  & \multirow{2}{*}{Re} & \multirow{2}{*}{Model} & Trainable& \multicolumn{2}{c}{MAE} \\
\cline{5-6}
 &  & &  Parameters & Velocity Loss & Pressure Loss    \\
\midrule
\multirow{2}{*}{Burgers'} & \multirow{2}{*}{$Re \in [1, 100] $} & \abrev & 75 K & 4 $\times$  $10^{-3}$ & -  \\
 & & PFNO & 466 K & 8 $\times$  $10^{-3}$ & -  \\
\midrule
\multirow{2}{*}{Incomp Navier-Stokes} & \multirow{2}{*}{$Re \in [1, 100] $} & \abrev & 140 K & 4 $\times$  $10^{-3}$ & 6 $\times$  $10^{-3}$  \\
 & & PFNO & 466 K & 0.01 & 0.012  \\
\bottomrule
\end{tabular}\\
\label{tab:model_uv}
\end{table}

\noindent Table \ref{tab:model_uv} highlights the clear advantage of \abrev~ over the PFNO when constrained to an identical data budget. For the Burgers' equation, \abrev~ achieves better accuracy in terms of velocity vector prediction. However for the INS test case, the superiority of our model is more clear in both velocity and pressure predictions. Notably, this performance gain is achieved with a considerable reduction in trainable parameters, highlighting the parameter efficiency and superior scaling of our modular, physics-informed architecture compared to traditional monolithic operators.
\subsection{Inference Speed}
A primary motivation for utilizing Scientific Foundation Models is the reduction of computational overhead associated with traditional numerical solvers. Table \ref{tab:model_speed} compares the time required to generate a single time step per trajectory across different physical tracks. Traditional simulations were performed on a uniform 128×128 Cartesian grid using the numerical schemes detailed in Appendix \ref{data_gen}. Benchmarking was conducted by comparing the traditional solver's execution on an AMD EPYC 7502P 32-Core Processor against the \abrev’s inference on a single NVIDIA A100.

\begin{table}[h]
\centering
\caption{Computational Time Comparison (Per time step)}
\begin{tabular}{lcccc}
\toprule
 & Convection-Diffusion & Scalar Burgers' & Vectorial Burgers' & INS \\
\midrule
Numerical Solver  & 0.064 s & 0.1 s & 0.22 s & 0.35 s\\
\midrule
\abrev & 0.0052 s & 0.0052 s & 0.0065 s & 0.01 s\\
\midrule
Speed-up & $\times$ 12 & $\times$ 19 & $\times$ 33 & $\times$ 35 \\
\bottomrule
\end{tabular}\\
\label{tab:model_speed}
\end{table}

\noindent For the scalar track, \abrev~ achieves a speed-up of 12$\times$ for Convection-Diffusion and 19$\times$ for the Viscous Burgers' equation. The efficiency gains are even more pronounced in the Vectorial Track, where \abrev~ achieves speed-ups of 33$\times$ for Vectorial Burgers' and 35$\times$ for the INS equations. The INS equations necessitate a multi-step Projection Method, involving the calculation of an intermediate velocity and the iterative solution of a global Pressure Poisson Equation. In contrast, \abrev~ bypasses these iterations through its modular architecture; by utilizing the pretrained Poisson Foundation Block and a single forward pass of the Physics-Informed Aggregator, the model enforces incompressibility and momentum transfer simultaneously. These results are effectively decoupled from the iterative complexity of the underlying physics.

\subsection{Discussion}
The evaluation of \abrev~ across the Convection-Diffusion, Burgers', and Incompressible Navier-Stokes equations showed promising results for the framework. The model acts as an aggregator that profits from the specialized knowledge transferred by the pretrained Foundation Blocks to solve new tasks. This confirms that the "Pre-train, Assemble, and Fine-tune" paradigm is a promising path for building scalable Scientific Foundation Models.\\

\begin{figure}[h]
    \centering
    \includegraphics[width=0.7\linewidth]{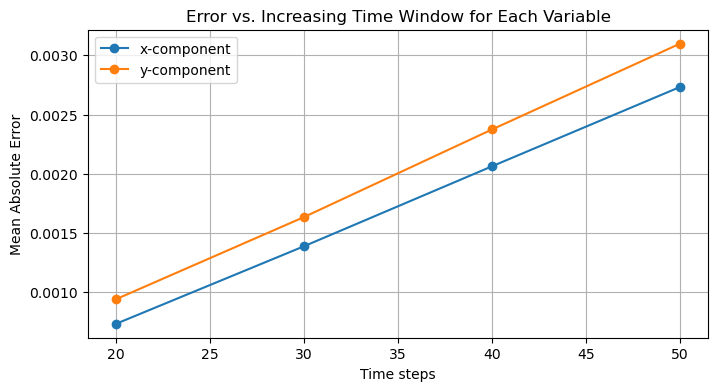}
    \caption{Comparative MAE for INS variables during the inference phase (T=50).}
    \label{fig:error_time}
\end{figure}

\noindent A critical observation in this 2D extension is the behavior of temporal error accumulation during autoregressive rollouts. As shown in Figure \ref{fig:error_time}, the error increases throughout the prediction sequence but remains within the same order of magnitude. This indicates that the model is numerically stable, as the error grows slowly and linearly, rather than exhibiting the exponential divergence as is often seen in monolithic solvers. By decomposing the system into modular pieces, the framework ensures that fundamental mechanisms are handled by experts who already understand those dynamics. This structural integrity prevents small temporal errors from collapsing into non-physical noise, keeping the overall solution physically valid even at the end of the rollout process. 

\section{Conclusion}
In this work, we have extended the \ourapp~ framework to two-dimensional systems, demonstrating a scalable and physically consistent approach to building Scientific Foundation Models. By leveraging the principle of compositionality, we decomposed complex fluid dynamics into a library of Foundation Blocks, each expert in a fundamental physical mechanism such as convection, diffusion, or nonlinear momentum transfer. These modules serve as a reusable base that can be assembled into task-specific solvers through an Adaptation Block. This ”pre-training assemble fine-tune” paradigm improves data efficiency, enhances interpretability, and enables compositional generalization to coupled physics not seen during pretraining.\\

\noindent \abrev~ demonstrates good performance on two-dimensional fluid dynamics systems such as convection–diffusion, Burgers', and INS equations. These results indicate that our approach is a promising way towards a scalable, physically consistent, and reusable foundation models. While the current operator library focuses on fluid flow equations, the proposed framework is inherently extensible. Additional physical dynamics can be incorporated by training new elementary operators, allowing the system to address new classes of problems without retraining the existing experts. Future work will concentrate on testing the model on more challenging configurations and geometries, and enriching the operator library to cover a broader range of PDEs. 

\bibliographystyle{ieeetr}
\bibliography{refrences}

\newpage

\appendix
\section{Architecture of Fourier Neural Operators}
\label{fno}
\subsection{The Fourier Neural Operator (FNO)}
The Fourier Neural Operator (FNO) \cite{li2020fourier} learns a mapping between functional spaces by parameterizing the integral kernel in the Fourier domain. Given an input $f(x)$, the FNO architecture consists of an initial lifting layer $P$ that maps the input to a high-dimensional representation $\varepsilon_0$, followed by $L$ Fourier layers, and a final projection layer $Q$ that maps the hidden representation to the output.

The core of the architecture is the Fourier Layer. For a hidden representation $\varepsilon_{\ell}$ at layer $\ell$, the update is defined as:
\begin{equation}
    \varepsilon_{\ell+1} = \sigma \left( \mathcal{F}^{-1} \left\{ \mathfrak{R}_{\ell} \left( \mathcal{F} \{ \varepsilon_{\ell} \} \right) \right\} + W_{\ell}(\varepsilon_{\ell}) \right),
\end{equation}
where $\epsilon_\ell \in \mathbb{R}^{d_h}$ denotes the feature embedding at layer $l$, $\sigma$ is a  nonlinear activation function, $\mathcal{F}$ and $\mathcal{F}^{-1}$ are the Fast Fourier Transform and the inverse one, $W_{\ell}$ performs a channel-wise linear transformation, and $\mathfrak{R}_{\ell}$ is a linear transformation function defined as:
\begin{equation}
    \mathfrak{R}_{\ell}(\mathcal{F}\{\varepsilon\})_{\kappa,i} = \sum_{j=1}^{d_h} (R_{\ell})_{\kappa,i,j} \mathcal{F}\{\varepsilon\}_{\kappa,j}, \\
\quad \kappa = 1, \ldots, \kappa^{\max} \quad \text{and} \quad i = 1, \ldots, d_h,
\end{equation} \\
with $R_{\ell}$ is a trainable weight tensor and $\kappa^{\max}$ is the maximum number of frequency modes.\\

\subsection{The Parametric Fourier Neural Operator (PFNO)}
While the standard FNO learns a global operator for a fixed physical system, the Parametric FNO (PFNO) \cite{yu2024parametric} generalizes across varying system parameters $\gamma$ (e.g., varying viscosity $\nu$ or velocity $\beta$).

\begin{figure}[h]
    \centering
    \includegraphics[width=0.8\linewidth]{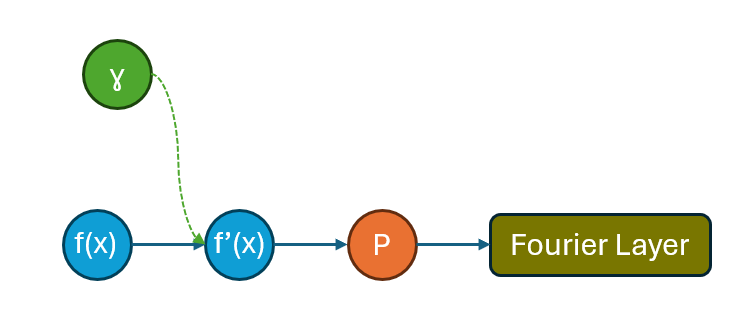}
    \caption{Architecture of the PFNO. Illustration of the parameter-dependent input transformation.}
    \label{fig:arch_pfno}
\end{figure}

As showed in Figure \ref{fig:arch_pfno}, PFNO incorporates parameter influence by appending each parameter value $\gamma \in  {\rm I\!R}^p$ to the codomain of the input function $f(x) \in  {\rm I\!R}^{d}$. The modified function $f'(x) \in  {\rm I\!R}^{d+p}$ is then projected to a higher dimension using the same lifting operator $P$.

\section{Data Generation and Numerical Schemes}\label{data_gen}
This appendix provides the technical details of the numerical methodologies used to generate the 2D datasets. All simulations were performed on uniform Cartesian grids with periodic boundary conditions, utilizing specific discretization schemes to ensure stability and physical accuracy.

\subsection{Elementary Operators (Foundation Blocks)}

The Foundation Blocks are trained on isolated physical mechanisms to learn disentangled representations of convection, diffusion, and nonlinear interactions.

\subsubsection{Scalar Foundation Blocks}
The \textbf{Scalar Velocity Track} library contains expert modules specialized in the fundamental transport mechanisms of a scalar field $u(x, y, t)$. Each block is pretrained on the following governing equations:

\begin{itemize}
    \item \textbf{Linear Convection Block:} Captures the transport of $u$ in a constant velocity field $\mathbf{\beta}$:
    \begin{equation}
        \frac{\partial u}{\partial t} + \beta \frac{\partial u}{\partial x} + \beta \frac{\partial u}{\partial y} = 0
    \end{equation}
    \item \textbf{Diffusion Block:} Learns the isotropic spreading of $u$ governed by the Laplacian operator and diffusion coefficient $\nu$:
    \begin{equation}
        \frac{\partial u}{\partial t} = \nu \left( \frac{\partial^2 u}{\partial x^2} + \frac{\partial^2 u}{\partial y^2} \right)
    \end{equation}
    \item \textbf{Nonlinear Convection Block:} Derived from the 2D inviscid Burgers' equation, this block models self-advection where the field $u$ acts as its own transport velocity:
    \begin{equation}
        \frac{\partial u}{\partial t} + u \frac{\partial u}{\partial x} + u \frac{\partial u}{\partial y} = 0
    \end{equation}
\end{itemize}

\subsubsection{Vectorial Foundation Blocks}
The \textbf{Vectorial Velocity Track} addresses the velocity vector field $\mathbf{u} = (u, v)$. These blocks are essential for assembling the Incompressible Navier-Stokes equations, with component-wise dynamics defined as:

\begin{itemize}
    \item \textbf{Vectorial Diffusion:} Applies the Laplacian operator independently to each velocity component:
    \begin{equation}
        \begin{cases}
            \frac{\partial u}{\partial t} = \nu \nabla^2 u \\
            \frac{\partial v}{\partial t} = \nu \nabla^2 v
        \end{cases}
    \end{equation}
    \item \textbf{Nonlinear Vector Convection:} Captures the momentum transfer where the vector field $\mathbf{u}$ advects itself:
    \begin{equation}
        \begin{cases}
            \frac{\partial u}{\partial t} + u \frac{\partial u}{\partial x} + v \frac{\partial u}{\partial y} = 0 \\
            \frac{\partial v}{\partial t} + u \frac{\partial v}{\partial x} + v \frac{\partial v}{\partial y} = 0
        \end{cases}
    \end{equation}
    \item \textbf{Poisson Foundation Block:} Solves the elliptic pressure-Poisson equation to relate the scalar pressure field $p$ to the velocity divergence source term $f$:
    \begin{equation}
        \nabla^2 p = \frac{\partial^2 p}{\partial x^2} + \frac{\partial^2 p}{\partial y^2} = f
    \end{equation}
\end{itemize}

\subsection{Coupled Physical Systems (Adaptation Blocks)}

The Adaptation Blocks are trained on complex systems where multiple elementary operators interact simultaneously.

\subsubsection{2D Convection-Diffusion}
This system couples linear transport with molecular dissipation:
\begin{equation}
    \frac{\partial u}{\partial t} + \mathbf{\beta} \cdot \nabla u = \nu \nabla^2 u
\end{equation}
We implement an \textbf{IMEX (Implicit-Explicit) scheme}. Convection is treated explicitly using a first-order upwind scheme for stability, while the diffusion term is treated implicitly via Crank-Nicolson. This hybrid approach ensures stability across a wide range of Péclet numbers (Pe) defined as:
\begin{equation*}
    \mathrm {Pe} = \dfrac {\beta \text{ x L}}{\nu}
\end{equation*}
where L is the characteristic length of the domain.
\subsubsection{2D Viscous Burgers' Equation}
The viscous Burgers' equation introduces coupling between nonlinear convection and diffusion:
\begin{equation}
    \frac{\partial \mathbf{u}}{\partial t} + (\mathbf{u} \cdot \nabla)\mathbf{u} = \nu \nabla^2 \mathbf{u}
\end{equation}
Data generation utilizes an \textbf{IMEX-RK3} approach. Nonlinear convection terms are computed explicitly via upwind discretization to prevent numerical oscillations, while viscous terms are handled implicitly. This equation is characterized by The Reynolds number (Re) defined as:
\begin{equation}
    \mathrm {Re} = \dfrac {U \text{ x L}}{\nu} ,
\end{equation}
where L is the characteristic length of the domain and $U$ is the characteristic value of $\mathbf{u}$. This dimensionless parameter help predict the behavior of fluids in the Burgers' equation as well as the Navier-Stokes equations.

\subsubsection{2D Incompressible Navier-Stokes}
The most complex coupled system involves momentum transport and the divergence-free constraint:
\begin{equation}
    \frac{\partial \mathbf{u}}{\partial t} + (\mathbf{u} \cdot \nabla)\mathbf{u} = -\frac{1}{\rho}\nabla p + \nu \nabla^2 \mathbf{u}, \quad \nabla \cdot \mathbf{u} = 0
\end{equation}
We implement a \textbf{Chorin-type Projection Method} consisting of three steps:
\begin{enumerate}
    \item \textbf{Tentative Velocity:} A velocity $\mathbf{u}^*$ is computed by solving the momentum equation (excluding pressure) via IMEX.
    \item \textbf{Pressure Poisson:} The Pressure Poisson Equation (PPE), $\nabla^2 p = \frac{\rho}{\Delta t} \nabla \cdot \mathbf{u}^*$, is solved to find the pressure field.
    \item \textbf{Correction:} The final velocity is updated via $\mathbf{u}_{t+1} = \mathbf{u}^* - \frac{\Delta t}{\rho} \nabla p$.
\end{enumerate}

\section{Closure Law}\label{closure}
\begin{itemize}
\item Convection-diffusion: Mathematically, let's denote $u$ the solution of the convection equation and $v$ the solution of diffusion equation. The aggregated output $w = \alpha_1 u + \alpha_2 v$ can be a solution of the convection-diffusion equation, where $\alpha_1$ and $\alpha_2$ represent mixing coefficients. By substituting the aggregated solution into the target equation:
\begin{equation}
     \frac{\partial (\alpha_1 u + \alpha_2 v)}{\partial t} + \beta \cdot \nabla (\alpha_1 u + \alpha_2 v) - \nu \cdot \Delta (\alpha_1 u + \alpha_2 v) = 0,
\end{equation}
that refers to :
\begin{equation}
     \alpha_1 (\frac{\partial u}{\partial t} + \beta \cdot \nabla u) + \alpha_2 (\frac{\partial v}{\partial t} - \nu \cdot \Delta v) 
     + \alpha_2 \beta \Delta v - \alpha_1 \nu  \nabla u
     = 0.
\end{equation}
We define $\mathcal{C}_{CD}$ as the residual of the convection-diffusion elementary operators:
\begin{equation}
    \mathcal{C}_{CD} = \alpha_1 (\frac{\partial u}{\partial t} + \beta \cdot \nabla u) + \alpha_2 (\frac{\partial v}{\partial t} - \nu \cdot \Delta v)  \approx 0
\end{equation}
and by simplifying $\mathcal{C}_{CD}$ as the knowledge transferred by the foundation blocks (equal to zero), we define close law of the convection-diffusion equation $\mathcal{C}_{cross}$ as: 
\begin{equation} 
     \mathcal{C}_{cross} = \alpha_2 \beta \nabla v - \alpha_1 \nu \Delta u 
     \label{optim1}
\end{equation}

\item Burgers' equation: Mathematically, let's denote $u$ as the solution of the inviscid Burgers' equation and $v$ as the solution from the diffusion equation. By substituting a point-wise linearized aggregation $w = \alpha_1 u + \alpha_2 v$ into the viscous Burgers' equation we obtain:
\begin{equation}
     \frac{\partial (\alpha_1 u + \alpha_2 v)}{\partial t} + (\alpha_1 u + \alpha_2 v) \cdot \nabla (\alpha_1 u + \alpha_2 v) - \nu \cdot \Delta (\alpha_1 u + \alpha_2 v) = 0,
\end{equation}
that refers to :
\begin{equation}
    \mathcal{C}_{B} + \alpha_1^2 u \cdot \nabla u + \alpha_2^2 v \cdot \nabla v + \alpha_1 \alpha_2 ( v \cdot \nabla u + u \cdot \nabla v) - \alpha_1 \nu  \cdot \Delta u = 0,
\end{equation}
where $\mathcal{C}_{B}$ is the residual of the Burgers' equation elementary operators:
\begin{equation}
    \mathcal{C}_{B} = \alpha_1 (\frac{\partial u}{\partial t} + u \cdot \nabla u) + \alpha_2 (\frac{\partial v}{\partial t} - \nu \cdot \Delta v) \approx 0,
\end{equation}
By simplifying $\mathcal{C}_{B}$, we define close law of the Burgers' equation $\mathcal{C}_{cross}$ as: :
\begin{equation}
    \mathcal{C}_{cross} = \alpha_1^2 u \cdot \nabla u + \alpha_2^2 v \cdot \nabla v + \alpha_1 \alpha_2 ( v \cdot \nabla u + u \cdot \nabla v) - \alpha_1 \nu  \cdot \Delta u
    \label{optim2}
\end{equation}

\end{itemize}

\end{document}